\documentclass[letterpaper, 10 pt, conference]{ieeeconf}  % Comment this line out if you need a4paper

\IEEEoverridecommandlockouts % This command is only needed if you want to use the \thanks command

\usepackage{graphics} % for pdf, bitmapped graphics files
\usepackage{epsfig} % for postscript graphics files
\usepackage{xcolor}
\usepackage{booktabs}
\usepackage{cite}
\usepackage{subcaption}
\usepackage{hyperref}

\usepackage{amssymb}  % assumes amsmath package installed

\title{\LARGE \bf
Language and Sketching: An LLM-driven Interactive Multimodal Multitask Robot Navigation Framework}

\author{Weiqin Zu$^{1}$, Wenbin Song$^{1}$, Ruiqing Chen$^{1}$, Ze Guo$^{2}$, Fanglei Sun$^{3}$$^{*}$, Zheng Tian$^{1}$, Wei Pan$^{4}$, and Jun Wang$^{5}$% <-this % stops a space
\thanks{$^{1}$ ShanghaiTech University, Shanghai, China} % {\tt\small \{zuwq2022, songwb, sunfl, zhengtian\}@shanghaitech.edu.cn}
\thanks{$^{2}$ Harbin Institute of Technology, Harbin, China}
\thanks{$^{3}$ University of Shanghai for Science and Technology, Shanghai, China}
\thanks{$^{4}$ The University of Manchester, Manchester, United Kingdom}
\thanks{$^{5}$ University College London, London, United Kingdom}
\thanks{$^{*}$Corresponding author. 
{\tt\small sunfanglei1@gmail.com}}
}

\begin{document}

\maketitle
\thispagestyle{empty}
\pagestyle{empty}

%%%%%%%%%%%%%%%%%%%%%%%%%%%%%%%%%%%%%%%%%%%%%%%%%%%%%%%%%%%%%%%%%%%%%%%%%%%%%%%%
\begin{abstract}
The socially-aware navigation system has evolved to adeptly avoid various obstacles while performing multiple tasks, such as point-to-point navigation, human-following, and -guiding.
However, a prominent gap persists: in Human-Robot Interaction (HRI), the procedure of communicating commands to robots demands intricate mathematical formulations.
Furthermore, the transition between tasks does not quite possess the intuitive control and user-centric interactivity that one would desire. 
In this work, we propose an LLM-driven interactive multimodal multitask robot navigation framework, termed LIM2N, to solve the above new challenge in the navigation field. 
We achieve this by first introducing a multimodal interaction framework where language and hand-drawn inputs can serve as navigation constraints and control objectives.
Next, a reinforcement learning agent is built to handle multiple tasks with the received information. 
Crucially, LIM2N creates smooth cooperation among the reasoning of multimodal input, multitask planning, and adaptation and processing of the intelligent sensing modules in the complicated system.
Detailed experiments are conducted in both simulation and the real world demonstrating that LIM2N has solid user needs understanding, alongside an enhanced interactive experience.
\end{abstract}
%%%%%%%%%%%%%%%%%%%%%%%%%%%%%%%%%%%%%%%%%%%%%%%%%%%%%%%%%%%%%%%%%%%%%%%%%%%%%%%%

\section{Introduction}
Mobile robots are becoming an integral part of our lives, used in domestic spaces and crowded places like train stations and airports.
In these places, robots need to comprehend the environment effectively and act in line with human intentions.
However, information obtained exclusively from laser scans is incomplete in complex environments.
For instance, there might be potential obstacles that the radar fails to detect. For example, a puddle of milk is spilled on the ground. Some obstacles require us to maintain a certain distance from them, like the surrounding area of a glass that one shouldn't touch or come near.
To address potential obstacles, while some work proposes the incorporation of visual data~\cite{majumdar2020improving,hong2021vln,lin2022adapt}, this demands both robust visual recognition and an understanding of obstacles, increasing the burden of system hardware and software. 
Therefore, we consider if these issues can be primarily addressed using 2D laser scans, supplemented with some low-dimensional, simple information.
In socially aware navigation, a robot's ability to provide services to users hinges on its understanding of complex environments. 
Understanding the environment is a basic requirement for point-to-point navigation; to enhance the user experience, robots should also provide a variety of additional services.
Although some studies\cite{morioka2004human,geetha2021follow,kastner2020deep} have achieved human-following and human-guiding tasks separately, and \cite{kastner2022human} have enabled a robot to perform both tasks, professionals still need to provide information through codes or mathematical formulations to assign and switch tasks, which is not a convenient way to interact for general users.
\begin{figure}[t]
  \centering
  \includegraphics[scale=0.15]{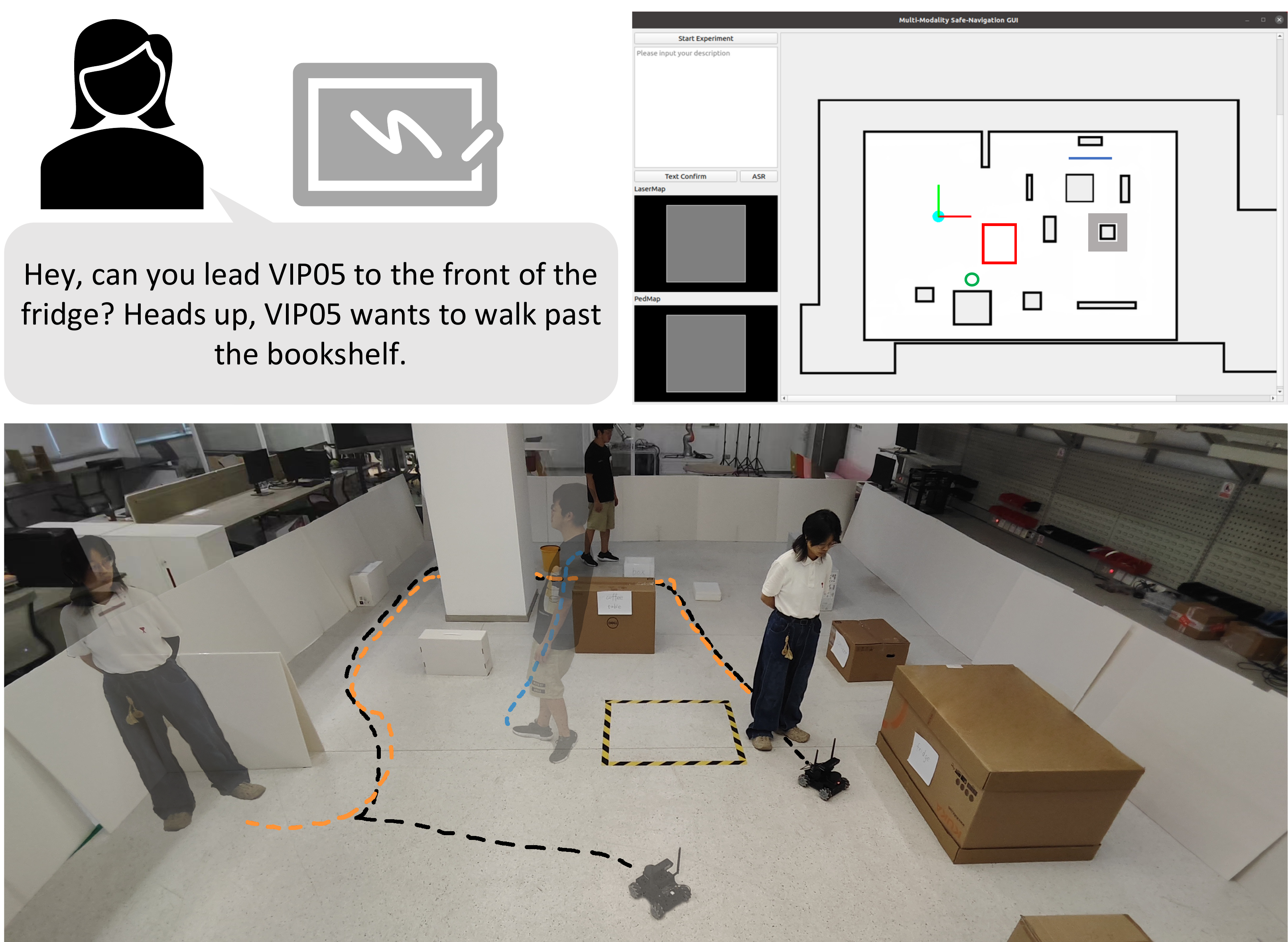}
  \caption{Users ask a robot to guide VIP05 to {\color{green} the fridge}, taking a route near the bookshelf ({\color{blue}{blue line}}). Challenges include an undetectable carpet area ({\color{red}{red box}}) and a box requiring distance ({\color{gray}gray box}). Users provide guidance through language and sketches input on an interactive interface.}
  \label{example_figure}
  \vspace{-10pt}
\end{figure}

A natural and direct interaction between humans and robots will be preferred as it can provide robots with critical information about the environment and greatly improve user experience. Hence, to enhance robots' understanding of the environment and free assignment of tasks, we consider a combination of language and hand-drawn sketches. In light of the advancements in Natural Language Processing (NLP),  Large Language Models (LLM) have been gaining powerful logical reasoning abilities. We envision robots comprehending human intentions from a single sentence input. Simultaneously, when certain details are difficult to describe verbally, users can supplement with simple sketches on an interactive interface. Therefore, we propose an LLM-driven Interactive Multimodal Multitask Robot Navigation Framework (LIM2N), a system capable of understanding multimodal user inputs from both language and sketches and executing various tasks. 
Users can provide robots with environmental information via text\footnote{By integrating the Automatic Speech Recognition (ASR) module, we enable voice input that is subsequently converted into text.} and supplementary sketches, as shown in Fig.~\ref{example_figure}.
Initially, an LLM module interprets textual input to determine the task type and environmental details, categorizing them into constraint and destination information.
In an intelligent sensing module, we process constraint information while simultaneously retrieving destination information.
In an RL module, we input the prior task type, destination information from the previous two modules, and constraint information processed by the intelligent sensing module to direct robots' movement.
We conduct experiments in both the simulation and the real world. The results indicate that LIM2N can more comprehensively interpret environmental information with enhanced stability. Additionally, user study feedback further validates our interaction model's intuitive user experience.

The main contributions of this work are the following:
\begin{itemize}
    \item We propose a novel navigation framework LIM2N based on language and sketches multimodal inputs in domestic and communal environments, significantly enhancing the user's interactive experience and the system's comprehensive understanding of the environment.
    \item We design a new algorithm where a moving robot, driven by LLM, completes multiple tasks mainly  based on adaptation and processing of the sensing modules, e.g., 2D laser scans, without the need to train multiple models. This differs from existing algorithms which often require multiple models for different tasks.
    \item We conduct both quantitative and qualitative experiments in the simulation and the real world to validate our framework's enhanced comprehension and interactivity.
\end{itemize}

\section{Related Work}
\textbf{LLM in navigation:} 
In mobile robot navigation tasks, providing textual directives for robots is an intuitive way to ensure robots understand user needs. Incorporating a language model with a visual module enhances performance in visual-linguistic navigation (VLN) tasks~\cite{hong2021vln}. Building on this, researchers have added an action prompt base to deepen robots’ environment comprehension~\cite{lin2022adapt}. Later, a transformer framework was developed to adjust navigation based on textual cues~\cite{bucker2023latte}. A linguistic-based interface for open-ended robotic control in human-machine interactions was presented~\cite{cui2023no}. However, challenges remain in textual understanding. The swift progress of LLM has spurred interest in robot linguistic control. Researchers have used the ChatGPT API to guide robots, presenting a novel approach for LLM in robotics~\cite{vemprala2023chatgpt}. A function library has been suggested to amplify LLM capabilities for complex tasks~\cite{liang2023code}, while others have focused on constraining LLM outputs to produce robot-executable directives~\cite{ahn2022can, lin2023text2motion}. Combining Visual Language Models with LLM, researchers have used value maps for robot path plotting~\cite{huang2023voxposer}, exemplifying efforts to assimilate multimodal data atop visual information~\cite{zhao2023chat}. Although text and visual tokens have been merged, the need for set icons limits its adaptability~\cite{jiang2022vima}. Given these methods’ reliance on visual data, a trade-off exists between environmental understanding accuracy and increased system loads, posing challenges for home robots and prompting researchers to explore non-visual environmental comprehension.

When relying solely on text, addressing indescribable information is challenging. Simple hand-drawn sketches may be the solution. Unlike visual data, these sketches, while less data-intensive, convey significant information. Sketches were used for vehicular navigation, enabling users to guide vehicles by drawing paths~\cite{skubic2003sketch}. This technique was further refined for single and multi-robot navigation~\cite{skubic2007using, boniardi2016autonomous}, and combined with laser data to improve robots’ self-localization~\cite{geetha2021follow}. Building on this, we explored using sketches to inform robots of potential obstacles in intricate environments.

\textbf{Multitasks navigation:} Understanding complex environments is crucial for delivering optimal services to users. Beyond basic point-to-point navigation, robots should engage in tasks like human-following and human-guiding. 
Given the unpredictability of pedestrian trajectories, many current methods utilize Deep Reinforcement Learning (DRL) techniques~\cite{yao2021crowd}. However, these approaches often fail to address the freezing robot problem. Expanding on this, researchers employ sound cues for smoother robot-pedestrian interactions~\cite{qiu2022learning}. Other works integrate pedestrian anticipated paths into the reward mechanism~\cite{jin2020mapless}. Although significant advancements have been made, many methods still focus primarily on point-to-point navigation. For better user experience, enhancing robot-pedestrian interactions is vital. Some studies integrated technologies like LIDAR for pedestrian tracking~\cite{chi2017gait}. When visuals are added, methods use pedestrian features for real-time follow-ups~\cite{algabri2020deep, gupta2016novel}.However, most current technologies require specialized knowledge and operate under certain constraints, neglecting intuitive user interaction. To address this issue, our research emphasizes user interaction, aiming to develop a system that accurately responds to direct user inputs.
% \section{Approach\wei{don't use approach or method, be concrete, use the exact algorithm or method name} \wei{no worries, you can use more than one sections, if you can't include everything in one section only}}
\section{LIM2N: An Interactive Framework}
\begin{figure*}[!htbp]
  \centering
  \includegraphics[width=\linewidth]{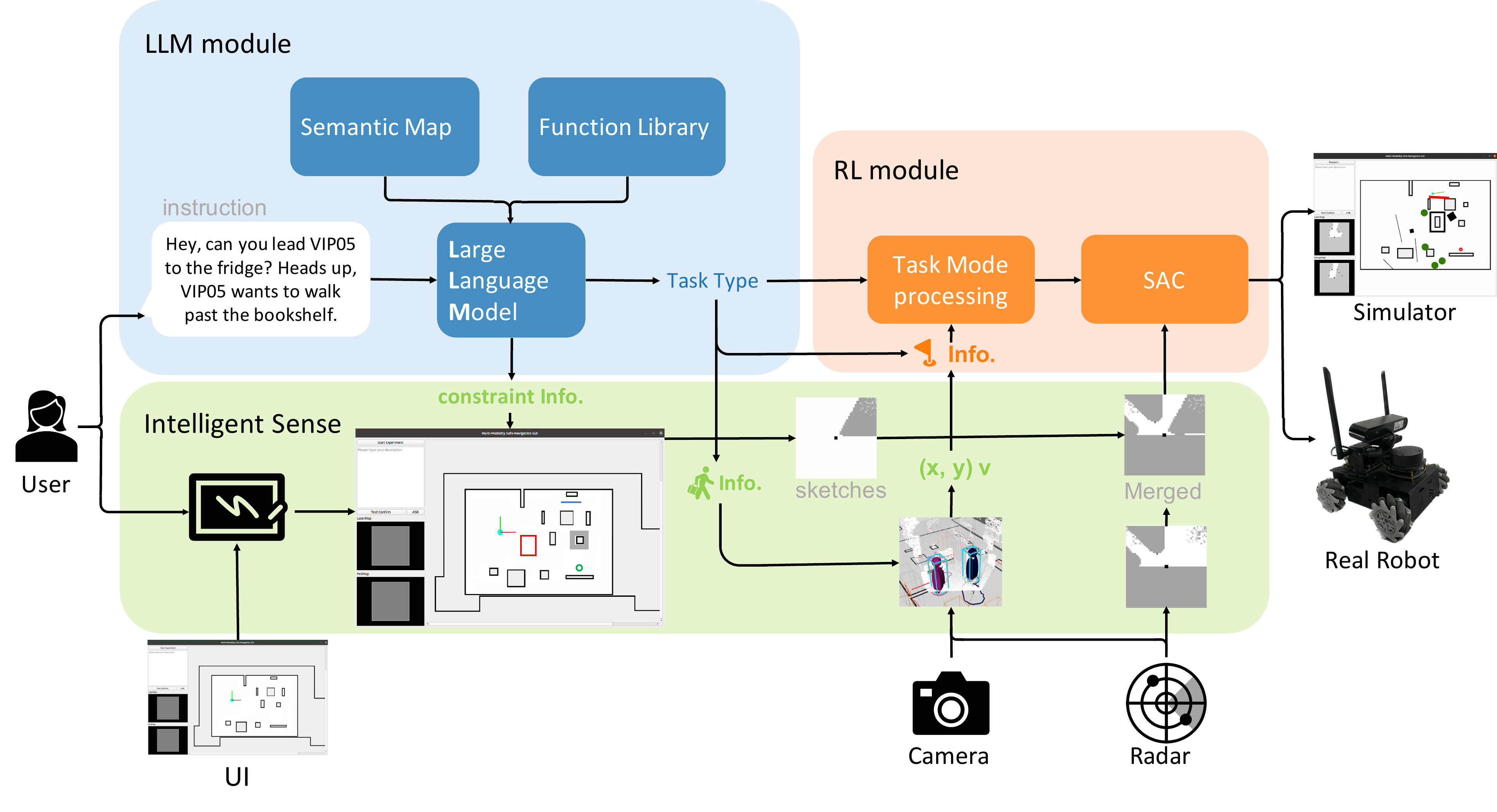}
  \caption{Overview of LIM2N. The framework contains an LLM module, an Intelligent Sensing Module, and a Reinforcement Learning Module.}
  \label{overview_figure}
  \vspace{-10pt}
\end{figure*}
The overview of our LIM2N is given in Fig.~\ref{overview_figure}. Our framework comprises an LLM module, an Intelligent Sensing module, and an RL module. Within the LLM module, input from text or voice is transformed into linguistic information and channeled into the LLM backbone. 
Utilizing a predefined text-based semantic map and function library, the LLM backbone identifies the necessary services a user needs, i.e. task type. From the text, it extracts and translates details into constraint information (e.g., potential obstacles) for the intelligent sensing module and destination information (e.g., end goal) for the RL module.
Within the intelligent sensing module, we integrate hand-drawn sketches received from users, constraint information from LLM module, and laser scans to produce a merged laser map.
Additionally, this module employs a robust pedestrian detection system SPENSER Spenser\footnote{\href{https://github.com/spencer-project/spencer\_people\_tracking}{https://github.com/spencer-project/spencer\_people\_tracking}} to provide the position and speed of pedestrians for multitask execution, with a specific emphasis on utilizing the camera exclusively for pedestrian detection within our work.
In the RL module, using task type and destination information, we pinpoint the robots' current target position. Together with the merged laser map, we then feed this into the Soft Actor-Critic (SAC).

\subsection{LLM module}
\begin{figure}[thpb]
  \centering
  \includegraphics[width=\linewidth]{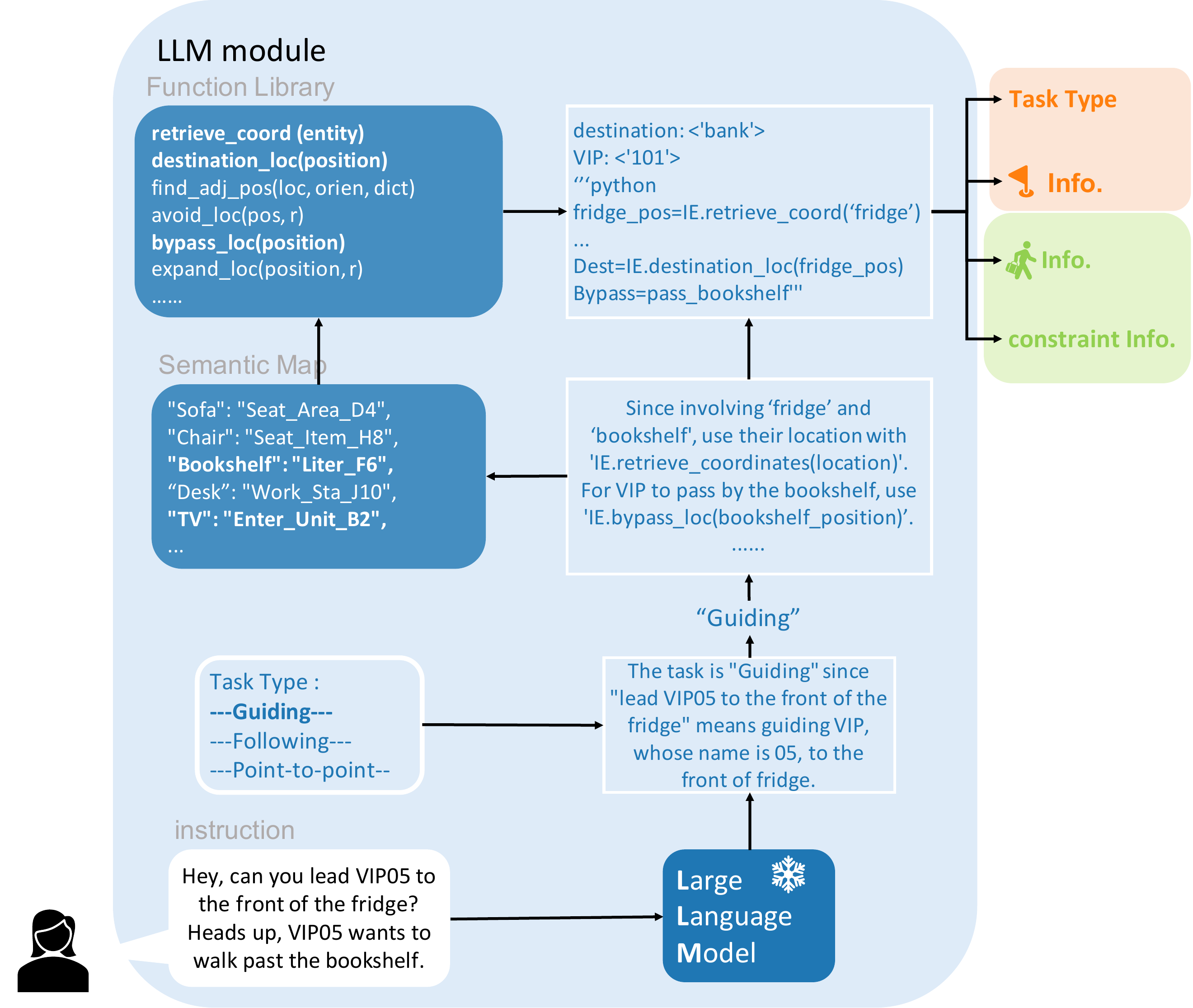}
  \caption{Processing an 'instruction' input through LLM backbone, determining a 'Guiding' service need. Using a semantic map, it identifies the fridge and the bookshelf coordinates. The function library sets the fridge as the end goal, and sketches a path via the bookshelf. Send the output to the RL module and the Intelligent Sense module, respectively.}  
  \label{LLM_figure}
  \vspace{-10pt}
\end{figure}

% \subsubsection{LLM Architecture}
Fig.~\ref{LLM_figure} shows the detail architecture of our LLM module. The LLM module accepts inputs either as text or voice. When the input is voice-based, it is converted to text by an ASR component.
% When textual input is received, it is understood by the LLM backbone.
Upon receiving textual input, the LLM backbone processes it.
Firstly, this backbone identifies the task type from a task-types pool.
Once the task requires end goal information, it will be treated as one type of destination information.
When tasks call for pedestrian data, like human-guiding, the pedestrian ID is integrated into the destination information and relayed to the intelligent sensing module for relevant data acquisition.
Beyond merely identifying tasks, LLM secondly extract environment-related details from the text. This includes destination information, for example taking routes that pass by a bookshelf before reaching a sofa, and constraint information concerning potential obstacles like carpets or restricted safety zones near glass cabinets.
We derive the necessary coordinates for environmental specifics from the semantic map. 
We process this information using predefined function libraries and third-party Python libraries like Numpy and Shapely. Subsequently, we send the destination information and the task type to the RL module, while assimilating the constraint information into the intelligent sensing module.
% interactive interface.

\subsubsection{Pre-configured Setup}
\textbf{Semantic map} is intended to compensate for the inability of LLM to directly interpret visual information.
Without visual cues, the LLM rely on linguistic inputs to understand their surroundings. Within LIM2N, a semantic map encompasses key building coordinates, as well as length and width, derived from 2D laser scans. 
\textbf{Function library} has been constructed to enhance the task completion capabilities of LLM and address their computational limitations. For example, the function \textit{expand\_location(position, r)} computes the coordinates based on \textit{position}, and subsequently determines the safety margin around it by considering its length, width, and the input parameter \textit{r}. Subsequently, we design a prompt for LLM, delineating the objectives to facilitate its comprehension of the high-level functions within the library. It is vital that LLM, upon encountering ambiguous instructions, actively seeks clarification from the user and refrains from making any unwarranted assumptions.

\subsection{Intelligent Sensing Module}
We developed a user interface (UI) to facilitate enhanced interaction with users for understanding the environment. Initially, we offer a map generated through post-processing of 2D laser scans of the real-world to allow intuitive user interaction. Users can sketch on the map as needed. These sketches appear within the UI, and users can select the end goal by clicking on the map. Users can also use sketches to convey wishes: drawing a closed pattern signifies a potential obstacle; tracing a route directs the robots' path; and outlining dangerous obstacles with extended lines indicates a safe distance to maintain.

Concurrently, we also provide text and voice input interfaces for the LLM modules in the UI. 
Once we retrieve the constraint information from the LLM module, it's directly visualized as sketches on the interface. These sketches, when amalgamated with the robot's 2D laser scans, form a merged laser map.
When the task requires pedestrian information, we use the robust \textit{Spenser} system to extract precise pedestrian data, notably the location $(x^t, y^t)$ and movement velocity $(v_x^t, v_y^t)$. 
The merged laser map and pedestrian data are both subsequently relayed to the RL module.

\subsection{Reinforcement Learning Module}
\begin{figure}[thpb]
  \centering
  \includegraphics[width=\linewidth]{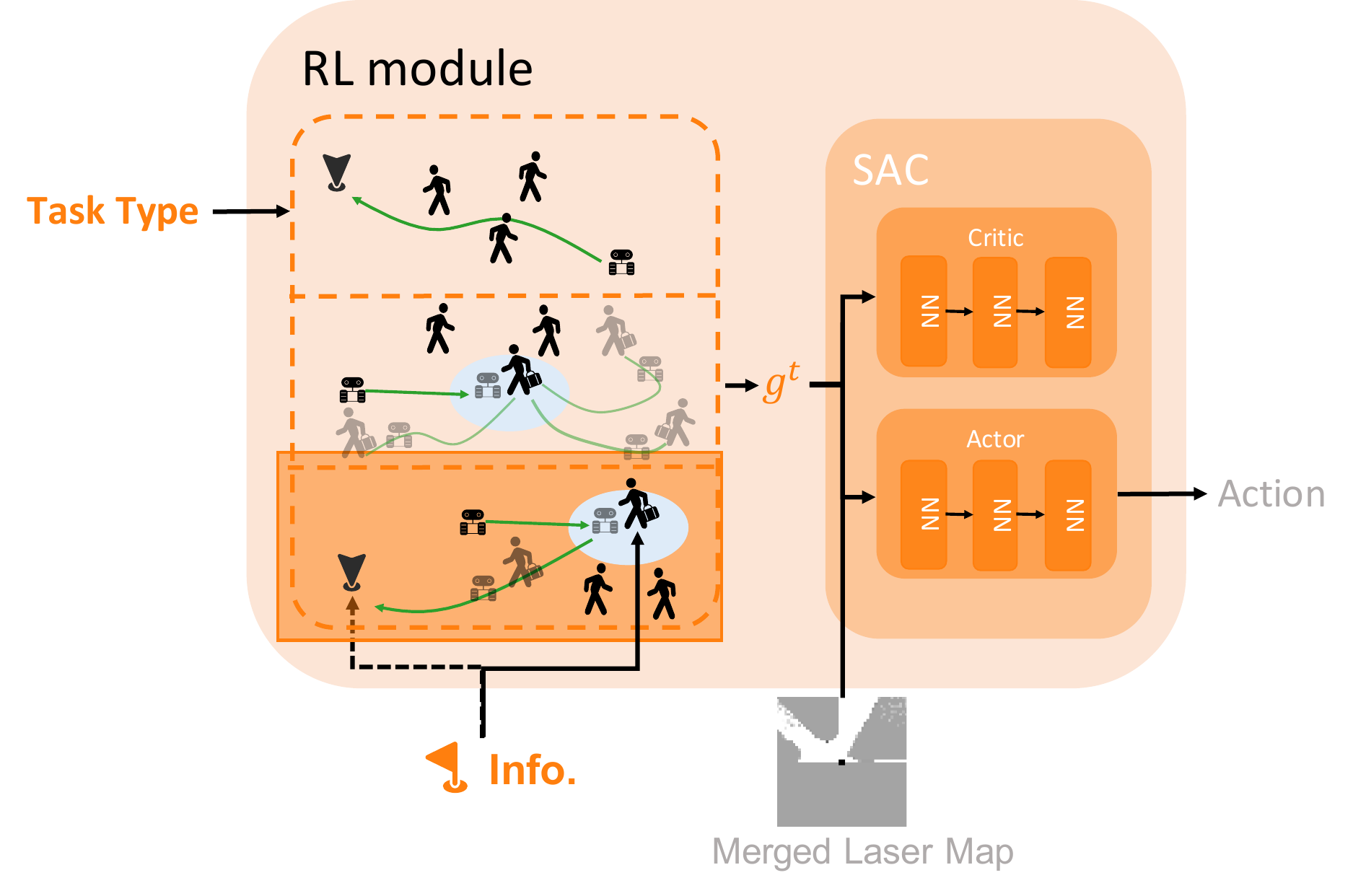}
  \caption{Using Task Mode processing, based on the task type and utilizing destination information (end goal and pedestrian positions), we determine the target location $g^t$ at time $t$. This $g_t$ and the merged laser map are provided to the SAC component as observations.}
  \label{RL_figure}
  \vspace{-10pt}
\end{figure}

Upon receiving specified data, we employ Task Mode Processing to determine the robots' target position $g^t$ for each task type at time $t$. Subsequently, both $g^t$ and the merged laser map are treated as observations for the subsequent processing. Utilizing the SAC algorithm, an advanced offline policy method, we deduce the robot's optimal action. See Fig.~\ref{RL_figure} for an overview.

\subsubsection{Task Mode Processing}
Referring to \cite{kastner2022human}, pedestrian assistance tasks in crowded environments include point-to-point navigation, human-following (where the robot follows a person), and human-guiding (where the robot leads a person to a specific location).
For point-to-point tasks, the robot must precisely know the end goal, and we directly designate the end goal as $g^t$.
During human-following, the robot only needs to continually locate the position of the person to be served (VIP) as $g^t$ and maintain a safe and proximate distance at all times.
In human-guiding, the robot initially sets $g^t$ to the VIP's location and approaches it. Once near the VIP, the robot updates $g^t$ to the end goal. Concurrently, the robot continuously monitors its distance from pedestrians. If the distance suddenly increases, perhaps due to unexpected pedestrian behavior or slow movement, the robot promptly adjusts $g^t$ to the pedestrian's position. After approaching the pedestrian, it then updates $g^t$ back to the end goal.

\subsubsection{SAC Neural Network Design}
We define the problem as a Partially-Observable Markov Decision Process (POMDP), represented by the tuple $< \mathcal{S}, \mathcal{O}, \mathcal{A}, \mathcal{P}, \mathcal{R} >$. Here $\mathcal{S}$ is the state space. $\mathcal{O}$ is the observation space.
% , which we define as a merged laser map data centered around the robot spanning 1.5m in four directions, data on nearby pedestrians, and the position of the target. 
$\mathcal{A}$ is the action space, where the action $a^t=(v^t, w^t)$ is continuous. 
% Here $v^t$ and $w^t$ represent the linear and angular velocities at time $t$, respectively. 
% Notably, as the real robots do not have a rear radar, it's logical for us to define $v^t$ as always positive. 
$\mathcal{P}$ is the state transition model $P(s'|s, a)$. $\mathcal{R}$ stands for the reward function. 
Inspired by \cite{yao2021crowd}, it includes elements like reward for reaching the target, collision penalty, distance from the target, and a negative reward for each step taken.

To train our navigation policy, we utilize a SAC algorithm, which mainly encompasses two components: an actor and a critic. The core objective of the training is to identify a policy $\pi$ that maximizes the accumulated reward, which is represented as:
$$\pi = \mathrm{argmax}_{\pi}{\sum_t\mathbb{E}_{(s^t, a^t)\sim \rho_{\pi}}[R(s^t, a^t)+\alpha H(\pi(\cdot|s^t))]},$$ where $H(\cdot | s^t)$ is the Entropy at state $s^t$, $\alpha$ is the relative importance of the entropy term against the reward.
Here, the value function is divided into the Q function and the V function, with the specifics of the formulas as follows: 
$$Q_{\pi}(s, a)=\mathbb{E}_{\rho_{\pi}}[\sum^{\infty}_{t = 0}\gamma^t R(s^t, a^t)+\alpha\sum^{\infty}_{t=1}\gamma^t H(\pi(\cdot|s^t))|s, a],$$ 
$$V_{\pi}(s) = \mathbb{E}_{\rho_{\pi}}[\sum^{\infty}_{t = 0}\gamma^t(R(s^t, a^t)+\alpha H(\pi(\cdot|s^t)))|s],$$
% $$Q_{\pi}(s, a)=\mathbb{E}_{\rho_{\pi}}[r(s^t, a^t)+\gamma^t V(\pi(s^t)).$$
where $\gamma$ is the discount factor in $(0,1)$. The update rule for the Q function is:
$$ Q_{\pi}(s, a) \leftarrow (1 - \alpha)Q_{\pi}(s, a) + \alpha^t(R + \gamma\mathbb{E}_{\rho_{\pi}}[V_{\pi}(s^t)] - \log(\pi))$$
% \begin{figure*}[thpb]
%     \centering
%     \begin{subfigure}[b]{0.325\textwidth}
%         \includegraphics[width=\linewidth]{P1.1.png}
%     \end{subfigure}
%     \hfill
%     \begin{subfigure}[b]{0.325\textwidth}
%         \includegraphics[width=\linewidth]{P1.2.png}
%     \end{subfigure}
%     \hfill
%     \begin{subfigure}[b]{0.325\textwidth}
%         \includegraphics[width=\linewidth]{P1.3.png}
%     \end{subfigure}
%     \\
%     \begin{subfigure}[b]{0.325\textwidth}
%         \includegraphics[width=\linewidth]{P1.4.png}
%     \end{subfigure}
%     \hfill
%     \begin{subfigure}[b]{0.325\textwidth}
%         \includegraphics[width=\linewidth]{P1.5.png}
%     \end{subfigure}
%     \hfill
%     \begin{subfigure}[b]{0.325\textwidth}
%         \includegraphics[width=\linewidth]{P1.6.png}
%     \end{subfigure}
%     \caption{In the simulator, tests increase in difficulty from 1 to 6. Gray: MC w/ IC trajectory; Blue: RL w/o IC (dotted lines indicate post-failure continuation); Orange: LIM2N trajectory.}
%     \label{path_figure}
% \end{figure*}

We trained our network within a simulated environment constructed using OpenCV\cite{qiu2022learning}. 
Within this environment, certain obstacles are fixed, and complemented by adding two randomly placed square obstacles and four line segments of different lengths in each training iteration. 
This dynamic obstacle configuration enhances the robot's proficiency in obstacle avoidance.
Additionally, leveraging pedestrian trajectories from SFM\cite{helbing1995social} and ORCA\cite{van2011reciprocal}, we incorporated four simulated pedestrians to ensure avoidance competence in a crowded environment.
\begin{figure}[htbp]
  \centering
  \includegraphics[width=\linewidth]{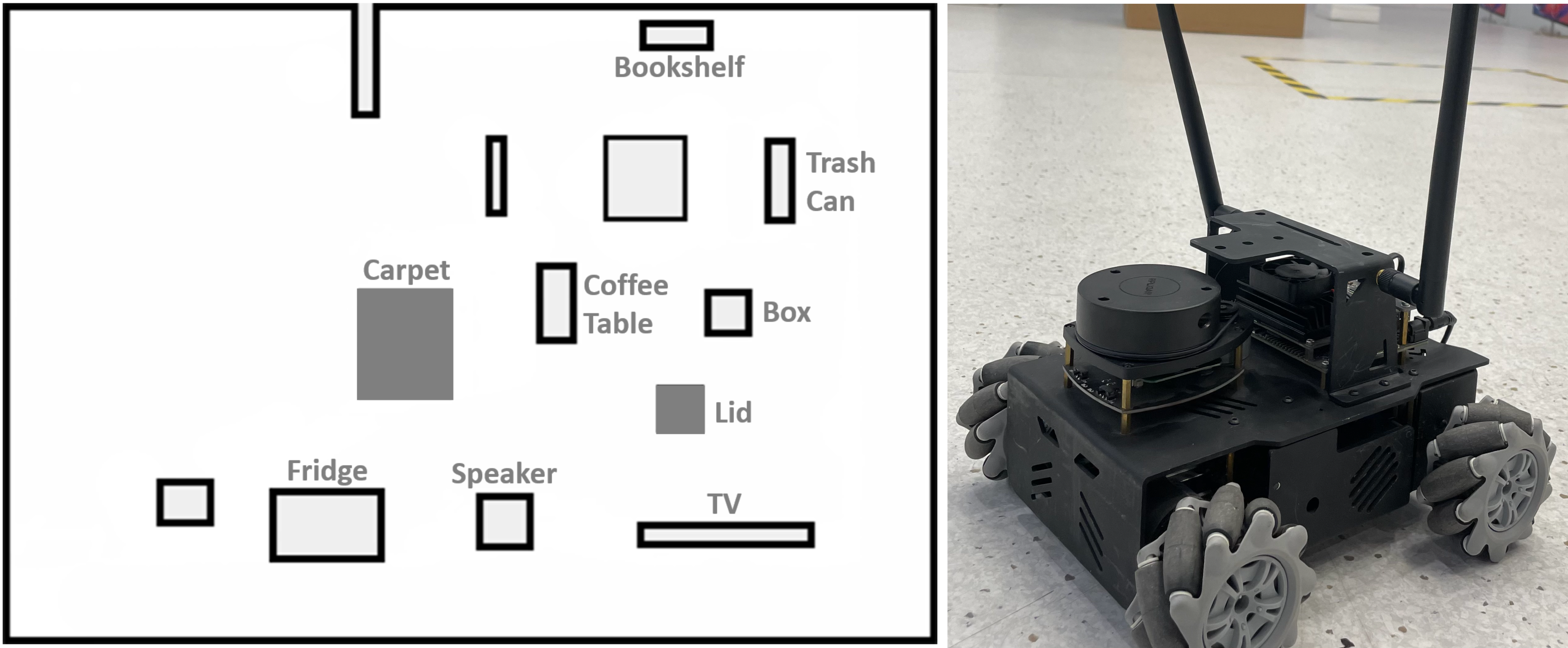}
  \caption{\textbf{Left}: Semantic map based on our simulator and real-world environment; \textbf{Right}: A robot in the real-world environment.}
  \label{map_figure}
  \vspace{-10pt}
\end{figure}
% \begin{figure}[thpb]
%     \centering
%     \begin{subfigure}[b]{0.25\textwidth}
%         \includegraphics[width=\textwidth]{simulation_env.pdf}
%         \label{env_simu}
%     \end{subfigure}
%     \begin{subfigure}[b]{0.22\textwidth}
%         \includegraphics[width=\textwidth]{robot.pdf}
%         \label{env_real}
%     \end{subfigure}
%     \caption{Experiments conducted under diverse settings in a simulator and a real-world environment.\zwq{change}}
%     \label{env_figure}
% \end{figure}

\section{Experiments}
We conducted experiments in both simulation and real-world to verify the performance of LIM2N. Our experiments aimed to: 1) 
evaluating the navigation performance of our framework with multimodal inputs, and 2) integrating language and sketches to improve the intuitiveness and convenience of interaction.
We measure the user-friendliness of interactions through language and sketches. For our real-world experiments, we utilized the HuanYuBot-H21X04 mobile robot, featuring the RPLIDAR A1 radar and a Mecanum wheel for motion, as depicted in Fig.~\ref{map_figure}.
\begin{figure}[thpb]
    \centering
    \begin{subfigure}[b]{0.15\textwidth}
        \includegraphics[width=\linewidth]{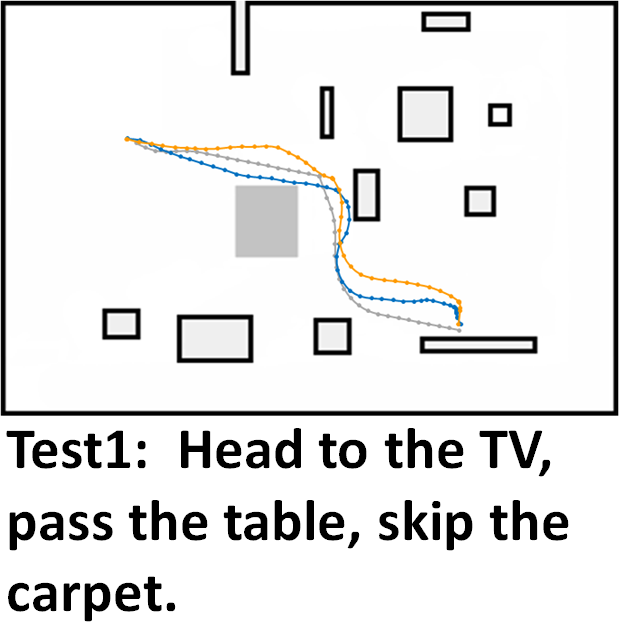}
    \end{subfigure}
    \hfill
    \begin{subfigure}[b]{0.15\textwidth}
        \includegraphics[width=\linewidth]{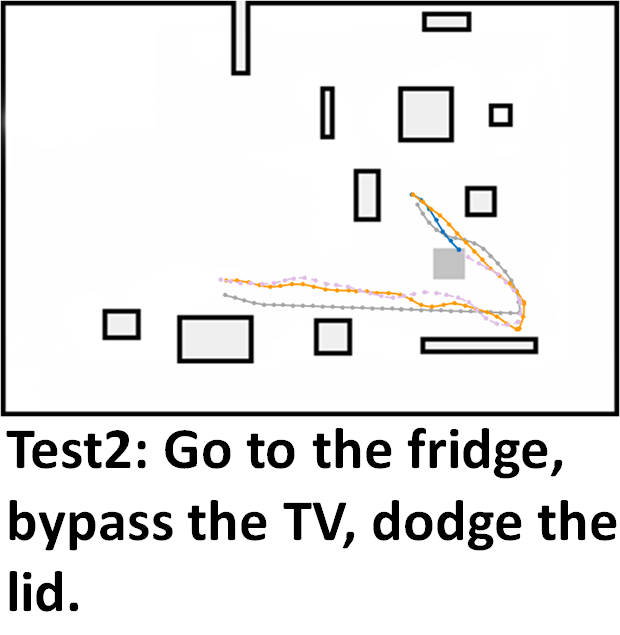}
    \end{subfigure}
    \hfill
    \begin{subfigure}[b]{0.15\textwidth}
        \includegraphics[width=\linewidth]{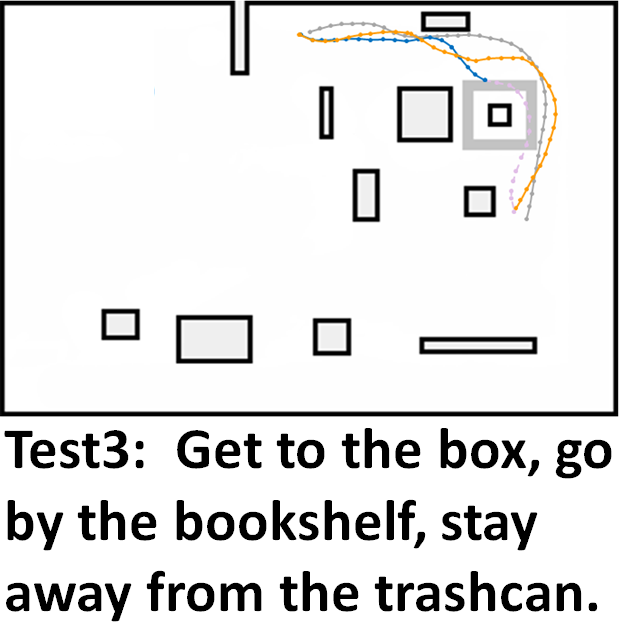}
    \end{subfigure}
    \\
    \begin{subfigure}[b]{0.15\textwidth}
        \includegraphics[width=\linewidth]{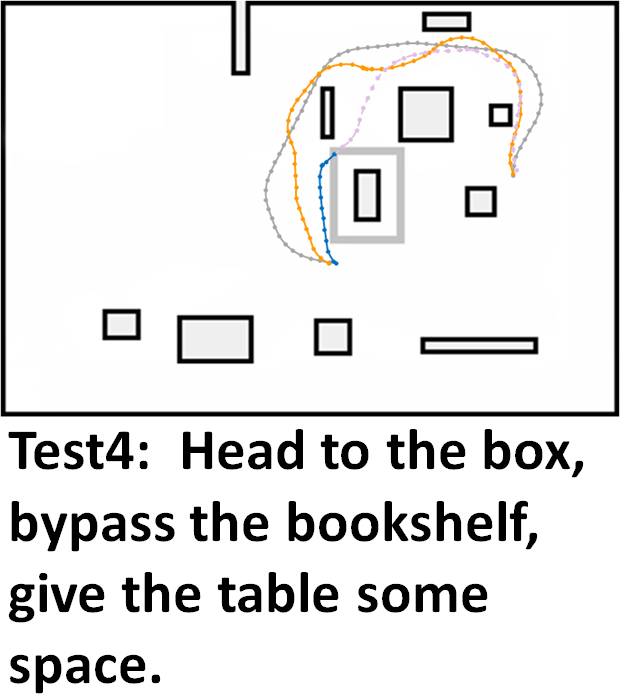}
    \end{subfigure}
    \hfill
    \begin{subfigure}[b]{0.15\textwidth}
        \includegraphics[width=\linewidth]{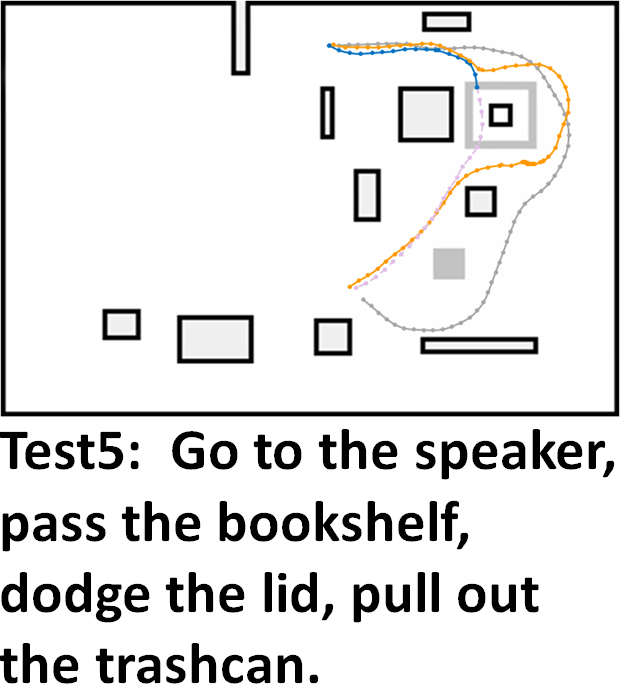}
    \end{subfigure}
    \hfill
    \begin{subfigure}[b]{0.15\textwidth}
        \includegraphics[width=\linewidth]{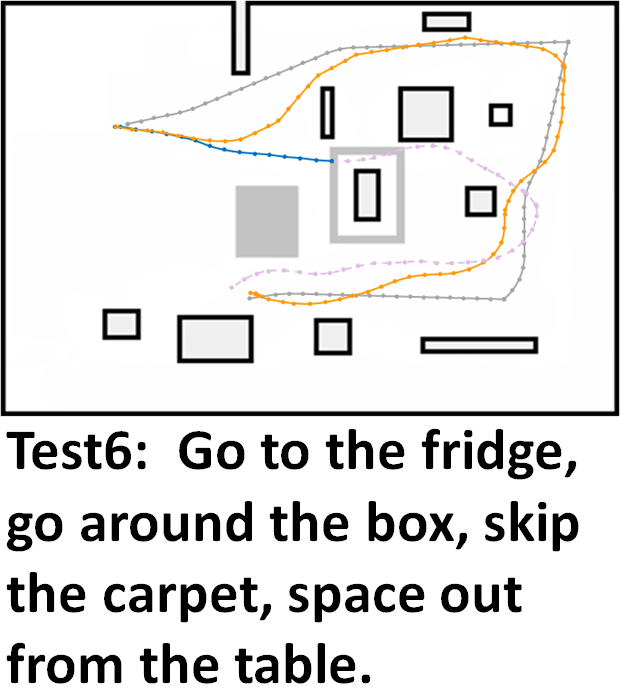}
    \end{subfigure}
    \caption{Navigation performance is evaluated using six tests. Gray: MC w/ IC trajectory; Blue: RL w/o IC (dashed lines indicate continuation after failure) trajectory; Orange: LIM2N trajectory.}
    \label{path_figure}
\end{figure}

\subsection{Navigation Performance}
\begin{figure}[thpb]
    \centering
    \begin{subfigure}[b]{0.23\textwidth}
        \includegraphics[scale=0.24]{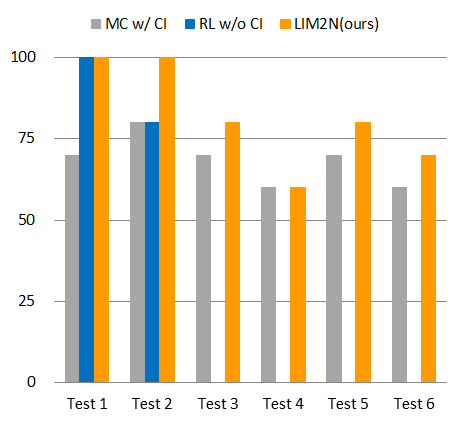}
        \caption{Static Scenario}
        \label{success_sta}
    \end{subfigure}
    %\hfill
    \begin{subfigure}[b]{0.23\textwidth}
        \includegraphics[scale=0.24]{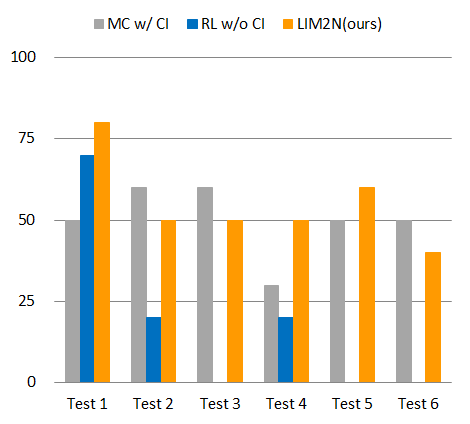}
        \caption{Pedestrian Scenario}
        \label{success_ped}
    \end{subfigure}
    \caption{The success rate comparison among MC w/ CI, RL w/o CI,  LIM2N in two simulation scenarios.}
    \label{success_figure}
    \vspace{-10pt}
\end{figure}

We constructed two simulation scenarios: the Static Scenario, based on a real scene with obstacles, and the Pedestrian Scenario, which builds upon the Static Scenario by featuring three moving pedestrians. 
Each scenario had six unique tests, and each test was repeated ten times specifically for that scenario, as detailed in Fig.~\ref{path_figure}. 
For the baseline, we used two approaches. The first approach, manual control with constraint information (MC w/ CI), involved ten participants operating the robot at LIM2N speeds after five initial tests. The second approach, RL without constraint information (RL w/o CI), allowed target setting only through code adjustments, without addressing potential obstacles. Success rates are shown in Fig.~\ref{success_figure}.
Additionally, TABLE~\ref{fail_table} categorizes the failure rates into three primary reasons: contacts with potential obstacles ($\alpha$), entries into safety zones ($\beta$), and failure to pass a specific point ($\gamma$). 

Our experiments demonstrate that LIM2N effectively navigates around potential obstacles, maintaining a stable speed and path. In manual control, operators interpret environmental cues, allowing them to circumvent potential obstacles. However, certain keyboard configurations might lead to inconsistencies in directional movement, causing confusion for operators. Inappropriately adjusted speed settings can either increase the risk of collisions or extend the test duration. The efficacy of manual control depends largely on the user's proficiency, leading to diverse outcomes. In contrast, the RL w/o CI encounters difficulties with potential obstacles, effectively addressing only visible ones. In our quantitative analysis, as shown in Fig.~\ref{success_sta}, LIM2N effectively processes environmental data, outperforming manual control in accuracy when navigating fixed obstacles. 
\begin{table}[htbp]
    \centering
    \caption{The failure rate in two simulation scenarios.}
    \begin{tabular}{ccccc}
         \toprule
          Environment&Methods&$\alpha$&$\beta$&$\gamma$\\ 
         \midrule
         &MC w/ CI&20\%&\textbf{17.5}\%&18.3\%\\
         Static Scenario&RL w/o CI&20\%&75\%&36.7\%\\
         &LIM2N(w/ CI)&\textbf{3.3\%}&22.5\%&\textbf{11.7\%}\\
         \midrule
         &MC w/ CI&43.3\%&\textbf{10\%}&38.3\%\\
         Pedestrian Scenario&RL w/o CI&36.7\%&82.5\%&45\%\\
         &LIM2N(w/ CI)&\textbf{26.7\%}&12.5\%&\textbf{23.3\%}\\
         \bottomrule
         \label{fail_table}
    \end{tabular}
    \vspace{-10pt}
\end{table}
However, its performance slightly lags behind manual control in the presence of virtual pedestrians, and current RL-based methods struggle to match human reflexes in emergencies with unpredictable obstacles, as illustrated in Fig.~\ref{success_ped}. RL w/o CI only succeeds in tests that do not require detecting potential obstacles or maintaining safe distances, failing when detours are necessary. 
In Table~\ref{fail_table}, the failure rates indicate that the primary shortcoming of RL w/o CI stems from breaches in the safety zone and collisions with unseen obstacles. 
Although manual control seldom enters safe zones, participants usually choose longer paths to ensure safety, as shown in Fig.~\ref{path_figure}. This also explains why, in individual cases, the success rate of manual control may exceed that of LIM2N.
Ultimately, LIM2N achieved notable results in both environmental comprehension and path planning.

\subsection{Interactive Navigation Efficacy}
In this study, we recruited 15 participants aged 20-27, with 5 females and 10 males, and 6 possessing a technical background. These participants had no prior involvement in this project.
% and lacked familiarity with our research topic. 
The experiments took place in a simulated living room, each lasting approximately 60 minutes. Participants controlled a mobile robot through three methods: 1) Manual control using a joystick; 2) Semi-automatic control requiring real-time updating of targets in the Rviz interface with RL w/o CI; 3) Our proposed method, combines text, voice, and sketches for test execution. To counter potential order effects, we randomized the experimental sequence and allotted time for participants to acclimate to the control methods.
\begin{figure}[thpb]
    \centering
    \begin{minipage}{0.28\textwidth}
        \begin{subfigure}[b]{\textwidth}
            \includegraphics[width=\linewidth]{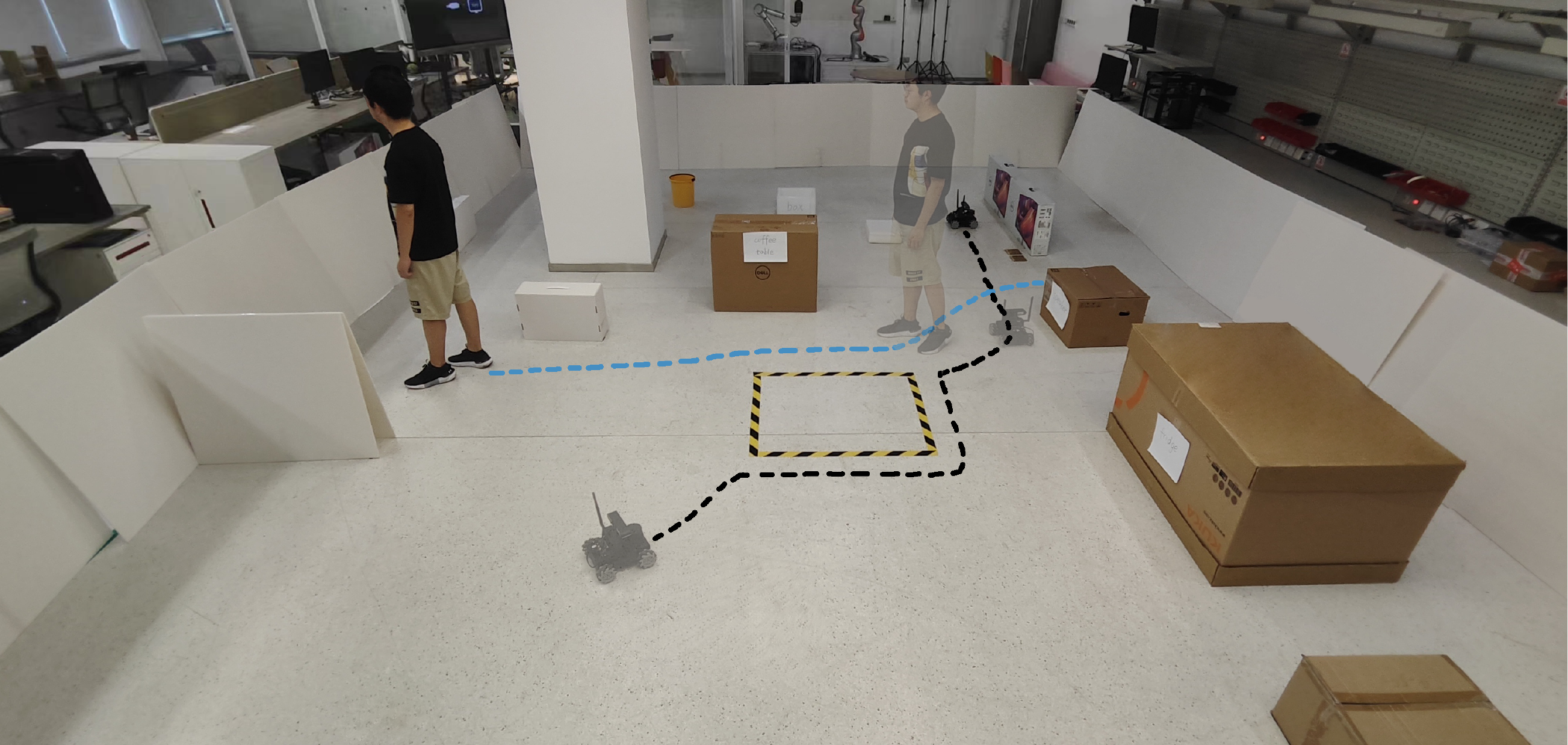}
        \end{subfigure}
        
        \vspace{6pt}

        \begin{subfigure}[b]{\textwidth}
            \includegraphics[width=\linewidth]{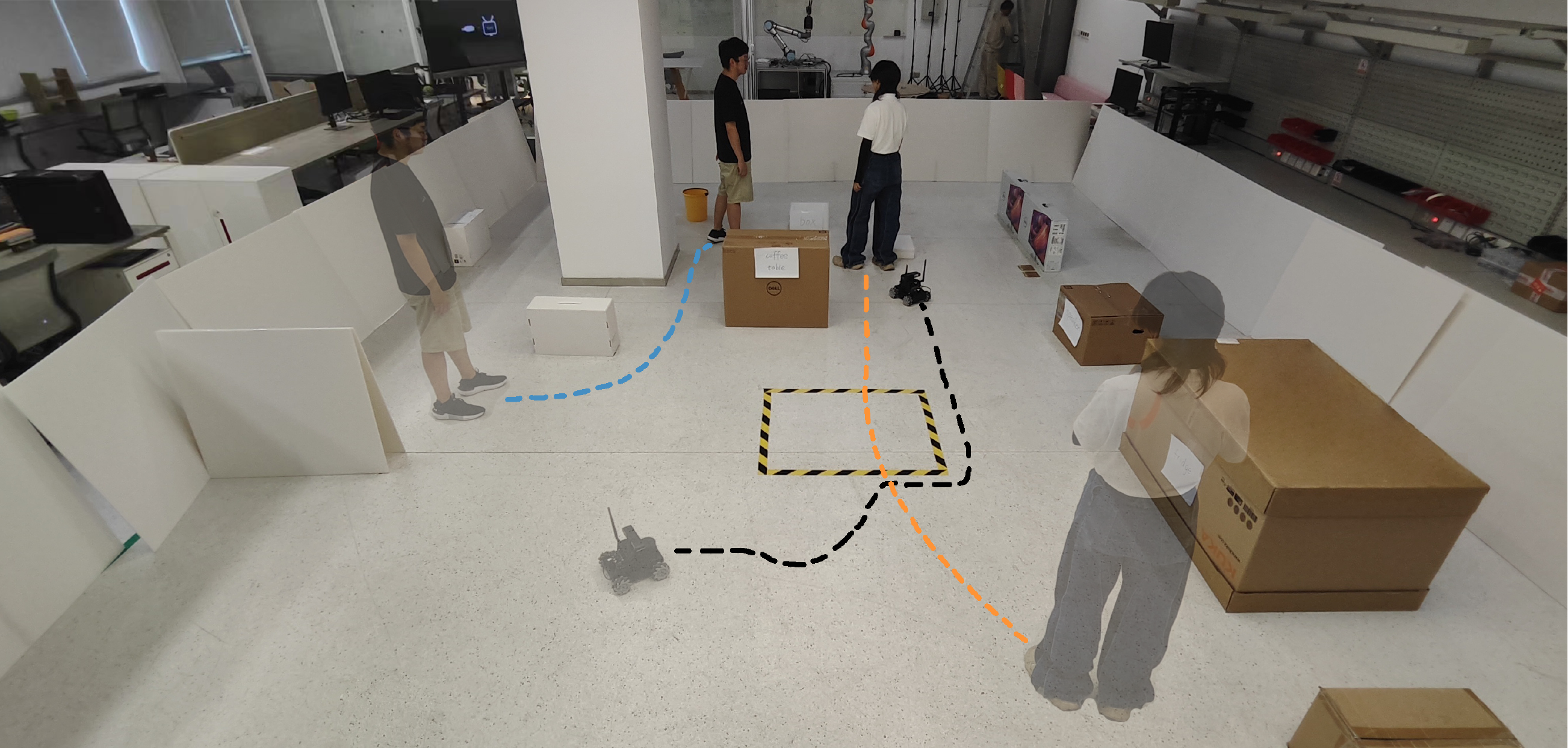}
        \end{subfigure}
        \label{inter_env_figure}
    \end{minipage}
    \hfill
    \begin{minipage}{0.19\textwidth}
        \includegraphics[width=\linewidth]{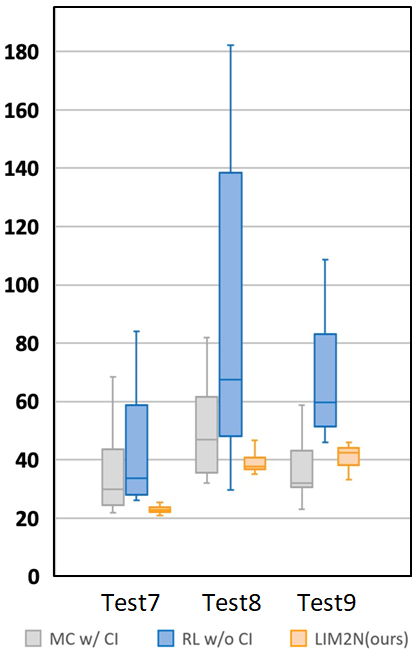}
        \label{finish_figure}
    \end{minipage}
    \caption{\textbf{Left}: The real-world experimental environment for user study.; \textbf{Right}: A box plot showing the mean values and the ranges of the three methods.}
    \label{inter_figure}
\end{figure}

In our research, based on prior tests, we incorporated human-following and human-guiding tasks, resulting in five tests, with some tests illustrated in Fig.~\ref{inter_figure}-Left. Some tests required participants to place obstacles undetectable to the robot to increase complexity. We recorded the completion times for three point-to-point navigation tests conducted by the participants, as shown in Fig.~\ref{inter_figure}-Right. LIM2N demonstrated significant robustness with minimal variance across users. Its average speed was comparable to other techniques, excelling in standard tests but lagging in more complex tests due to differences between simulated RL algorithms and real-world robot dynamics. Based on this, LIM2N proves to be both more efficient and stable.
\begin{figure}[thpb]
  \centering
  \includegraphics[width=\linewidth]{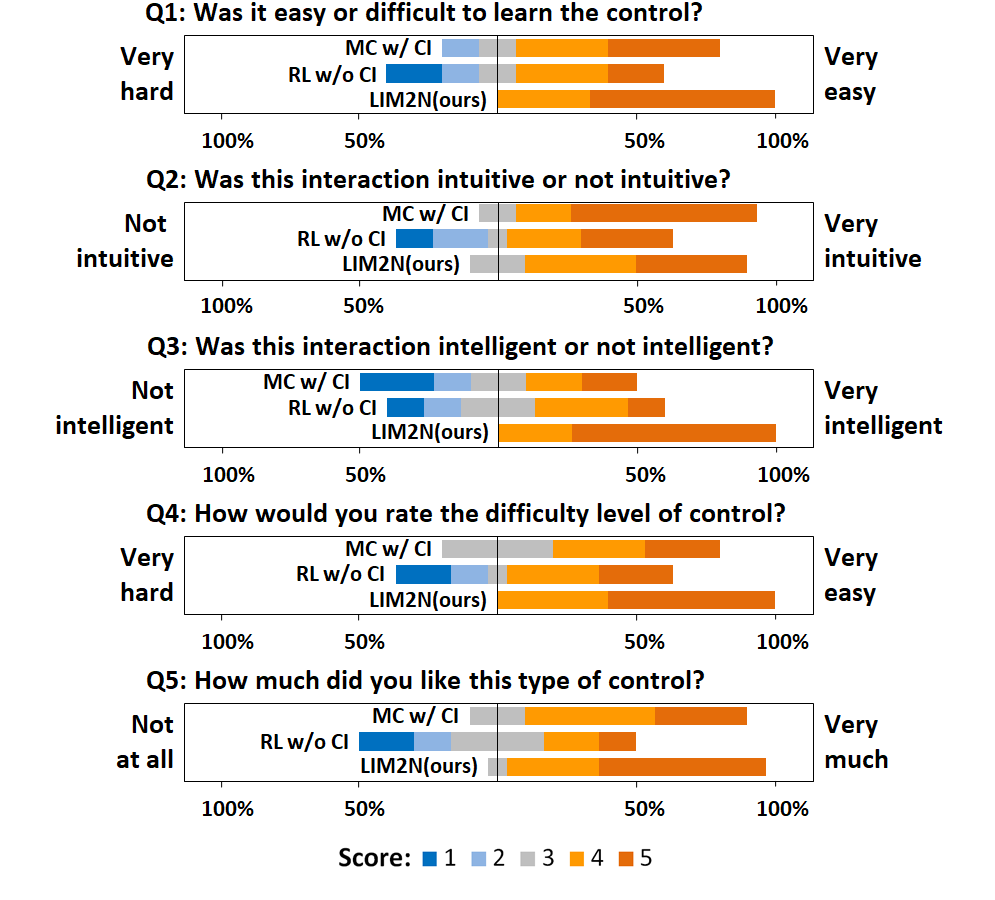}
  \caption{Bars left of center represent negative feedback, while those to the right signify positive responses.}
  \label{inter_5_figure}
\end{figure}

After the experiments, participants rated their experiences using a 5-point Likert scale, summarized in Fig.~\ref{inter_5_figure}. The results showed LIM2N has made advancements in enhanced learning, intelligent operation, and improved control. LIM2N's intuitiveness and user preferences were similar to joystick controls. Conversations with participants favoring joystick control revealed their preference for manual interaction, influenced by their affinity for mechanics. Many noted that on the Rviz platform, updating based on the real-time status of the robot was not as convenient as using a joystick due to the inability to see potential obstacles. 
Overall, LIM2N provided a satisfactory interactive experience, highlighting the enhanced user experience it provides.

\section{Conclusion}
We have introduced an LLM-driven multimodal input framework that conveys environmental information to robots via text or hand-drawn inputs. This approach also enables the robot to execute diverse tests based on users' specific needs. 
Experimental results from both simulated and real-world environments demonstrate that LIM2N exhibits a profound ability to comprehend environmental information while also boasting robust interactive capabilities.
In future iterations of this work, we plan to: 1) incorporate human feedback to realize a human-in-the-loop system, utilizing user feedback as learning data for the RL algorithm, and 2) address the challenge of managing multi-user information when multiple users seek services from the robot.

% %%%%%%%%%%%%%%%%%%%%%%%%%%%%%%%%%%%%%%%%%%%%%%%%%%%%%%%%%%%%%%%%%%%%%%%%%%%%%%%%

\bibliographystyle{IEEEtran}
\bibliography{root}

\end{document}